\documentclass[11pt]{article}
\usepackage{eamt23}
\usepackage{times}
\usepackage{url}
\usepackage{latexsym}
\usepackage{amsmath}
\usepackage[small,bf]{caption} 
\title{Beyond Semantics: Measuring Fine-Grained Emotion Preservation in Small Language Model-Based Machine Translation}

\author{Dawid Wiśniewski\\
  Poznań University of Technology\\
  {\tt dawid.wisniewski@cs.put.poznan.pl}  \And
  Igor Czudy\\
  Poznań University of Technology\\
  }

\date{}

\begin{document}
\maketitle
\begin{abstract}
Preserving affective nuance remains a challenge in Machine Translation (MT), where semantic equivalence often takes precedence over emotional fidelity. This paper evaluates the performance of three state-of-the-art Small Language Models (SLMs) -- EuroLLM, Aya Expanse, and Gemma -- in maintaining fine-grained emotions during backtranslation. Using the GoEmotions dataset, which comprises Reddit comments across 28 distinct categories, we assess emotional preservation across five European languages: German, French, Spanish, Italian, and Polish. Specifically, we investigate (i) the inherent capability of these SLMs to retain emotional sentiment, (ii) the efficacy of emotion-aware prompting in improving preservation, and (iii) the performance of ModernBERT as a contemporary alternative to BERT for emotion classification in MT evaluation.
\end{abstract}

\section{Introduction}
The primary objective of Machine Translation (MT) has traditionally been the preservation of semantic equivalence, ensuring that the \textit{what} of a message is accurately conveyed across linguistic boundaries. However, as Large Language Models (LLMs) increasingly mediate human communication through chatbots, virtual assistants, and localized content, the \textit{how} of a message, its emotional resonance and affective nuance -- has become equally critical.


Despite the rapid progress in neural MT, preserving fine-grained emotions remained a formidable challenge~\cite{DBLP:conf/amta/LoharAW18}. Languages encode affect through diverse linguistic mechanisms, ranging from specific lexical choices and modal particles to complex syntactic shifts. These nuances are often lost in translation pipelines that prioritize word-for-word or purely semantic accuracy.


The current landscape of Natural Language Processing is witnessing a shift toward Large Language Models (LLMs) and, increasingly, Small Language Models (SLMs) designed for localized deployment. While massive models like Gemini or GPT dominate general benchmarks, SLMs (typically under 10 billion parameters) offer sustainable, efficient, and often more specialized alternatives for regional or task-oriented applications~\cite{van2025survey}. In the European context, the emergence of smaller models like \textbf{EuroLLM}~\cite{martins2025eurollm}, optimized for EU languages, presents a unique opportunity to study affective transfer in a multi-family linguistic environment (Slavic, Germanic, and Romance). Simultaneously, models like \textbf{Aya Expanse}~\cite{dang2024aya}, developed with a focus on massive multilingual instruction tuning, represent a more global-focused approach to cross-lingual understanding.

In this paper, we evaluate the ability of \textbf{EuroLLM}~\cite{martins2025eurollm}, \textbf{Aya Expanse}~\cite{dang2024aya}, and Google's \textbf{Gemma}~\cite{team2024gemma} to transfer emotional content from English into German, French, Polish, Spanish, and Italian followed by backtranslation to English. We leverage the GoEmotions dataset~\cite{demszky2020goemotions}, which provides a high-resolution taxonomy of 28 distinct categories (27 emotion labels and a neutral class). This allows us to move beyond binary sentiment (positive/negative), into the territory of complex human states such as remorse, pride, and curiosity.

Given that LLM outputs are highly sensitive to input configurations, a key component of this research is investigating whether emotion-aware prompting, providing explicit instructions to prioritize emotional preservation, significantly enhances model performance. Finally, to evaluate the stability of emotional signals in backtranslated text, we benchmark contemporary classifiers, comparing traditional encoders like BERT~\cite{DBLP:conf/naacl/DevlinCLT19} and DeBERTa v3~\cite{he2021debertav3} against the recently introduced \textbf{ModernBERT}~\cite{warner2025smarter}.

Our research is guided by the following core research questions:
\begin{itemize}
\item \textbf{RQ1}: How do various SLM architectures compare in their ability to preserve fine-grained emotions during the translation process?
\item \textbf{RQ2}: Which specific emotional categories are most susceptible to degradation during translation?
\item \textbf{RQ3}: To what extent does explicit emotion-aware prompting improve the emotional fidelity of SLM-generated translations?
\item \textbf{RQ4}: Does ModernBERT offer better classification quality for detecting emotional content compared to established encoder models like BERT and DeBERTa?
\end{itemize}





\section{Related work}
\paragraph{Emotion preservation in MT}

The preservation of emotional fidelity is a significant area of inquiry within Machine Translation (MT). Recent literature has explored this challenge through various lenses, ranging from early neural architectures to modern generative models.

Early investigations into affective loss often utilized backtranslation as a diagnostic tool. \cite{troiano2020lost} evaluated emotional signal degradation in English-to-German and English-to-Russian pipelines. While they observed a notable loss of affective information, their study was limited to WMT'19 FAIRSEQ models~\cite{ott2019fairseq}, GloVe embedding-based classification~\cite{pennington2014glove}, and seven core emotions (anger, disgust, fear, guilt, joy, sadness, and shame), leaving the performance of modern autoregressive models on a wider emotional spectrum largely unexplored. Similarly, \cite{kajava2020emotion} examined the viability of annotation projection, noting that while sentiment is generally preserved, more nuanced emotional information is frequently lost due to incomplete translations or lexical ambiguity in the target language.




More recent studies have highlighted the persistent difficulty of maintaining nuance in commercial systems. \cite{qian-etal-2023-evaluation} demonstrated that Google Translate failed to preserve original emotions in over 50\% of English-to-Chinese translations, identifying polysemous words and negations as primary error drivers. To mitigate losses, \cite{brazier2024conditioning} explored multimodal approaches, enriching LLM inputs with features from Speech Emotion Recognition (SER) models~\cite{wagner2023dawn}. Their findings suggest that conditioning translations on affective variables, particularly arousal, yields measurable gains in translation quality.

The recent dominance of Large Language Models (LLMs), such as Google’s Gemini~\cite{team2023gemini}, Anthropic’s Claude~\cite{DBLP:journals/iswa/CaruccioCPSST24}, and OpenAI’s GPT-4o~\cite{singh2025openai}, has shifted the MT paradigm. Findings from the Workshop on Machine Translation (WMT 2025) confirm that these general-purpose models often surpass specialized NMT systems; notably, Gemini 2.5 Pro achieved state-of-the-art results across multiple language pairs despite not being exclusively trained for translation~\cite{kocmi2025findings}.



The specific capacity of Small Language Models (SLMs) to manage fine-grained affective transfer -- especially when guided by emotion-aware prompting -- remains a critical research gap that this paper aims to address.





\paragraph{Small language models \& MT}

The trajectory of Large Language Model (LLM) research has recently bifurcated, with significant momentum shifting toward Small Language Models (SLMs) optimized for edge deployment and computational efficiency. This paradigm shift is driven by the realization that high-quality data curation and advanced distillation techniques enable models with significantly fewer parameters to rival the performance of massive architectures while remaining compatible with consumer-grade hardware. This accessibility facilitates localized fine-tuning and greater granular control over model behavior~\cite{van2025survey}.

Recent releases have defined the current SLM landscape. Meta’s Llama 3.2 family~\cite{DBLP:journals/corr/abs-2407-21783} introduced ultra-lightweight 1B and 3B variants specifically designed for mobile and edge platforms. These models utilize structural pruning and knowledge distillation from larger Llama counterparts to maintain sophisticated reasoning capabilities within a compact footprint. Similarly, Google’s Gemma 2 series~\cite{team2024gemma} and particularly the 2B and 9B versions leverage a distillation-heavy training objective to achieve an exceptional performance-to-parameter ratio, often outperforming much larger models on general reasoning benchmarks.

Multilingual accessibility has been a primary driver for regional SLM development. The EuroLLM project~\cite{martins2025eurollm} focuses on providing native support for 35 languages, with a specific emphasis on the official languages of the European Union. By balancing data mixtures across diverse linguistic datasets, \textbf{EuroLLM} versions effectively mitigate the English-centric bias prevalent in earlier architectures. In a similar vein, Cohere’s Aya Expanse (8B and 32B)~\cite{dang2024aya} employs multilingual arbitrage, preference training, and model merging to deliver state-of-the-art performance across 23 languages. Furthermore, DeepSeek-V3~\cite{liu2025deepseek} represents an effort to harmonize high computational efficiency with superior agentic performance through innovations such as DeepSeek Sparse Attention (DSA) and a large-scale agentic task synthesis pipeline.

Beyond general-purpose reasoning, these SLMs have demonstrated high proficiency in specialized linguistic tasks, including high-fidelity machine translation~\cite{song2025small} and automated grammatical error correction~\cite{wisniewski2025exploring}. To facilitate the practical deployment of these models on resource-constrained hardware, quantization remains essential. Activation-aware Weight Quantization (AWQ)~\cite{DBLP:journals/sigmobile/LinTTYXH24} has emerged as a preferred method for compressing SLMs to 4-bit precision. Unlike traditional uniform quantization, AWQ identifies salient weights -- those corresponding to high-magnitude activations, and protects them from aggressive quantization errors, thereby preserving model performance. Other widely adopted formats, such as GGUF, BitsAndBytes, and GPTQ~\cite{rajput2024benchmarking}, provide additional trade-offs between universal compatibility and raw inference throughput.

\paragraph{Emotion taxonomies and datasets}
The computational modeling of emotion in text is fundamentally grounded in two primary psychological frameworks: categorical and dimensional. Historically, Ekman’s model~\cite{ekman1992argument} has served as the foundational categorical baseline, identifying six universal emotions: anger, disgust, fear, happiness, sadness, and surprise. While this simplicity is advantageous for broad sentiment analysis tasks, it often lacks the granularity required to capture the complexities of human expression. Conversely, Plutchik’s model~\cite{plutchik1980general} proposes a hierarchical "wheel" of emotions that accounts for varying affective intensities and polarities. Complementing these are dimensional frameworks, such as the Valence-Arousal-Dominance (VAD) model~\cite{russell1977evidence}, which represent emotional states as continuous coordinates in a multi-dimensional vector space rather than as discrete labels.

The operationalization of these theories within the Natural Language Processing (NLP) domain has been facilitated by a diverse array of benchmark datasets. Early efforts, such as the International Survey on Emotion Antecedents and Reactions (ISEAR)~\cite{scherer1990international}, provide high-quality, self-reported affective descriptions across seven categories; however, its limited scale poses challenges for training modern deep learning architectures. To capture the informal and dynamic nature of contemporary digital communication, datasets like SemEval-2018 Task 1 (Affect in Tweets)~\cite{mohammad-etal-2018-semeval} introduced multi-label emotion intensity tasks, enabling models to predict co-occurring affective states. For dimensional analysis, EmoBank~\cite{buechel-hahn-2017-emobank} remains a critical resource, mapping 10,000 sentences directly to the VAD space.

Recently, the GoEmotions dataset~\cite{demszky2020goemotions} has emerged as a good choice for fine-grained affective classification. By labeling 58,000 Reddit comments across 27 distinct emotional categories accompanied by a neutral class, it allows Small Language Models (SLMs) to distinguish between subtle emotional states, such as remorse versus grief or admiration versus love, that remained conflated in traditional 6-class models. This high-resolution taxonomy is particularly suited for evaluating the affective bleaching or shifts that may occur during the machine translation process.

\section{Dataset}
We utilize the GoEmotions dataset~\cite{demszky2020goemotions}, a manually annotated corpus comprising 57,732 English Reddit comments. The dataset employs a fine-grained taxonomy of 27 emotion labels plus a neutral class. Each instance was reviewed by 3 to 5 annotators, with additional metadata identifying examples deemed ambiguous by the experts.

\paragraph{Dataset preprocessing}
To optimize the dataset for training robust emotion classifiers, we implemented a filtering and refinement procedure:

\begin{enumerate}
\item \textbf{Noise Reduction:} We excluded all examples marked as "ambiguous" by at least one annotator to ensure high-confidence ground truth labels.
\item \textbf{Label Refinement:} We removed the \textit{neutral} class to focus the study specifically on active emotional transfer. For the remaining 27 categories, we applied a consensus-based merging strategy, retaining only those labels identified by at least two annotators.
\item \textbf{Class Balancing:} To mitigate severe class imbalance, where frequent labels like \textit{admiration} outnumber sparse labels like \textit{grief} by a factor of 30, we removed categories with a representation lower than one standard deviation below the mean class frequency.
\item \textbf{Stratified Splitting:} The resulting data was partitioned into training (80\%) and test (20\%) sets. We employed the Iterative Stratification algorithm~\cite{sechidis2011stratification} to ensure that the complex multi-label emotion distribution remained consistent across both subsets.
\end{enumerate}

\begin{figure*}
    \centering
    \includegraphics[width=1\linewidth]{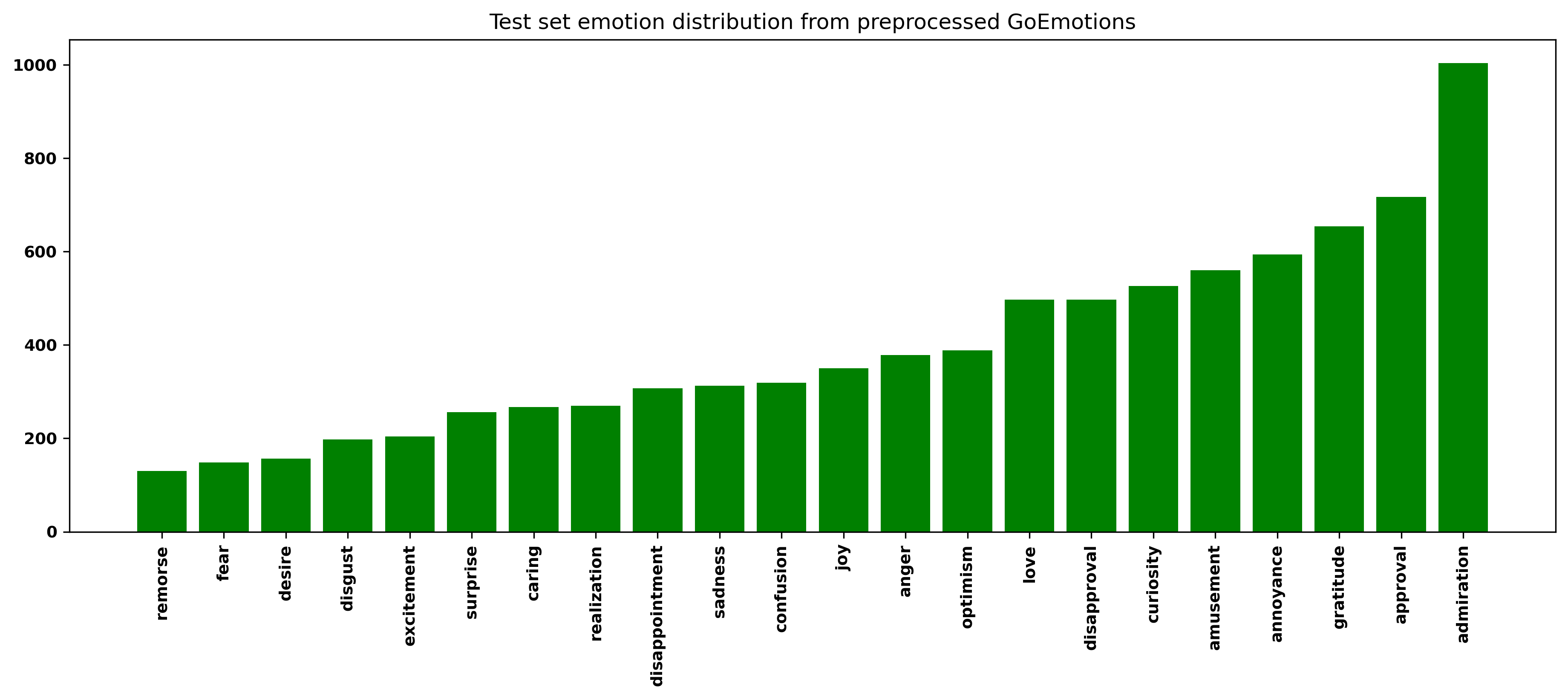}
    \caption{Emotions distribution in the preprocessed testset, which represents 20\% of the full dataset.}
    \label{fig:preprocessed_test}
\end{figure*}

After preprocessing, the final dataset consisted of 29,544 training and 7,386 test examples across 22 emotion categories. Five emotions (\textit{pride, relief, grief, embarrassment, and nervousness}) were excluded due to insufficient sample sizes. The final test set distribution is visualized in Figure~\ref{fig:preprocessed_test}.





\section{Methodology \& Experimental setup}






Our experimental framework is designed to measure how effectively Small Language Models (SLMs) preserve the 22 identified emotions through a round-trip translation process.

\subsection{Backtranslation Pipeline}

We perform backtranslation (English → Target → English) across five pivot languages: German, French, Spanish, Italian, and Polish. For the translation and backtranslation tasks, we compare three state-of-the-art SLMs:
\begin{itemize}
\item \textbf{EuroLLM-9B-Instruct-AWQ}\\~\cite{martins2025eurollm}~\footnote{\url{https://huggingface.co/stelterlab/EuroLLM-9B-Instruct-AWQ}}
\item \textbf{Aya-Expanse-8B-AWQ}\\~\cite{dang2024aya}~\footnote{\url{https://huggingface.co/Orion-zhen/aya-expanse-8b-AWQ}}
\item \textbf{Gemma-2-9B-IT-AWQ}\\~\cite{team2024gemma}~\footnote{\url{https://huggingface.co/solidrust/gemma-2-9b-it-AWQ}}
\end{itemize}
To accommodate hardware constraints (24,GB VRAM), all models were deployed using 4-bit Activation-aware Weight Quantization (AWQ). We utilized the vLLM engine~\cite{10.1145/3600006.3613165} with greedy decoding (temperature = 0) to ensure deterministic and reproducible outputs.

\subsection{Prompt Engineering}

To test the impact of instruction sensitivity on affective preservation, we evaluated two zero-shot prompt configurations:

\begin{enumerate}
    \item \textbf{Basic ($P_{base}$):} \textit{Translate the following text from LANG1 to LANG2.$\backslash$n$\backslash$n TEXT} 
    \item \textbf{Emotion-Aware ($P_{emo}$):} \textit{Translate the following text from LANG1 to LANG2. Please focus on preserving the emotions, tone, and intensity of the original text.$\backslash$n$\backslash$n TEXT} 

\end{enumerate}

Since models often append comments to their responses when using complex prompts, we search for double newlines, a common separator for such additions, and trim them as well as the content provided after those markers.

This setup yields a comprehensive evaluation matrix of 30 backtranslated test set variants (3 models × 5 languages × 2 prompts).

\subsection{Emotion Classification as Evaluation}

To quantify emotional loss, we fine-tune three encoder architectures the original English training set:

\begin{itemize}
    \item \texttt{BERT base cased}\\~\cite{DBLP:conf/naacl/DevlinCLT19}~\footnote{\url{https://huggingface.co/google-bert/bert-base-cased}}
    \item \texttt{DeBERTA-v3 base}\\~\cite{he2021debertav3}~\footnote{\url{https://huggingface.co/microsoft/deberta-v3-base}}
    \item \texttt{ModernBERT base}\\~\cite{warner2025smarter}~\footnote{\url{https://huggingface.co/answerdotai/ModernBERT-base}}
\end{itemize}

These models serve as our ground truth classifiers. We then evaluate their performance on the backtranslated variants, interpreting any drop in $F_1$-score relative to the original test set as evidence of affective degradation.

Those models are of similar size, with BERT, DeBERTa, and ModernBERT having 110M, 184M, and 149M parameters, respectively.

The models were trained with the same hyperparameters: 
\begin{itemize}
\item \texttt{num\_train\_epochs=10},
\item \texttt{problem\_type} set to \texttt{multi\_label\_classification},
\item \texttt{batch\_size=32}.
\end{itemize}
We applied early stopping with patience set to 1, evaluation strategy set to epoch, and micro-$F_1$ score for early stopping/best model selection. 

\subsection{Evaluation metrics}
\label{sec:metrics}
Let:
\begin{itemize}
    \item $D$ be the original English test set with gold-standard labels.
    \item $L$ be the set of pivot languages ($\{\text{German, French, Spanish, Italian, Polish}\}$).
    \item $BT(D, M, \pi, \ell)$ represent the backtranslated version of $D$ generated by model $M$ using prompt $\pi$ via pivot language $\ell \in L$.
    \item $\Phi$ be a fixed emotion classifier trained on the training set.
    \item $F_1(\Phi, X)$ denote the macro-averaged $F_1$ score of classifier $\Phi$ on dataset $X$ relative to the original gold labels.
\end{itemize}

The \textit{Affective Drop} $\Delta_{aff}$ for a specific model-prompt-classifier configuration $(M, \pi, \Phi)$ measures the loss of $F_1$ score averaged over all languages as compared to the reference score and is defined as:

\begin{equation}
\small
\Delta_{\text{aff}}(M, \pi, \Phi) =  F_1(\Phi, D) - \frac{1}{|L|} \sum_{\ell \in L} F_1(\Phi, BT(D, M, \pi, \ell))
\label{eq:affective_drop}
\end{equation}

Interpretation:
\begin{itemize}
    \item $\Delta_{\text{aff}} \approx 0$: Indicates near-perfect preservation of affective signals; the translation process did not significantly alter the emotional features recognized by the classifier.
    \item $\Delta_{\text{aff}}>0$: Represents a loss of emotional fidelity, where higher values indicate greater "affective bleaching" or the introduction of emotional noise during translation.
\end{itemize}

We can also introduce a per-emotion affective drop $\Delta^{e}_{aff}$ calculated for the emotion $e$ as follows: 

\begin{equation}
\small
\Delta_{\text{aff}}^{e}(M, \pi, \Phi) = F_1^e(\Phi, D) - \frac{1}{|L|} \sum_{\ell \in L} F_1^e(\Phi, BT(D, M, \pi, \ell))
\label{eq:per_emotion_drop}
\end{equation}
, where $F_1^e$ is the $F_1$-score for the class $e$.

\section{Results and Analysis}

\subsection{Aggregate Performance and Classifier Sensitivity}
The results of our backtranslation experiments are summarized in Table~\ref{tab:average-drops}. Each entry represents the average affective drop ($\Delta_{\text{aff}}$) for a specific combination of SLM, emotion classifier, and prompt type. As $\Delta_{\text{aff}}$ calculates the average drop over languages and emotion categories, it gives a general perspective on SLMs, classifiers, and prompts.

Overall, the analyzed models exhibit robust performance in preserving affective nuance, with average $\Delta_{\text{aff}}$ scores ranging from 2.89 to 4.93 percentage points. Despite these relatively low drops, we observe distinct performance hierarchies across both the translation models and the evaluation classifiers.

\subsection{Comparative Analysis of SLMs and Classifiers}
Our analysis reveals a consistent performance gradient among the tested models. Specifically:
\begin{itemize}
\item \textbf{SLM Hierarchy:} \textbf{EuroLLM} consistently outperforms both Aya and Gemma across all configurations. When comparing results using identical prompts and classifiers, we observe a stable ranking: \\
EuroLLM $>$ Aya $>$ Gemma \\The best-performing configuration overall, the one achieving the lowest affective drop utilizes EuroLLM with the emotion-aware prompt ($P_{emo}$) evaluated via ModernBERT.
\item \textbf{Classifier Robustness:} There is a visible difference in how the underlying encoder architectures perceive affective degradation. \textbf{ModernBERT} consistently yields the lowest $\Delta_{\text{aff}}$, followed by DeBERTa-v3 and BERT. This may suggest that ModernBERT’s architecture is better suited for detecting fine-grained emotional consistency in MT evaluation.
\end{itemize}

\subsection{The Impact of Emotion-Aware Prompting}
One of the most notable findings of this study is the marginal impact of explicit emotional instructions on backtranslation quality.

\paragraph{Classification Robustness:}
As shown in Table~\ref{table:emo_drop}, the introduction of $P_{emo}$
  has a negligible effect on the final $F_1$ scores. While EuroLLM shows a slight, consistent improvement (a smaller drop) when using $P_{emo}$, the gains are statistically marginal. For Aya and Gemma, the basic prompt ($P_{base}$
 ) actually yielded superior results. This suggests that contemporary SLMs inherently prioritize emotional preservation during translation, and explicit prompting may, in some cases, introduce noise or over-correction.

\paragraph{Semantic Translation Quality:}
To ensure that emotional preservation does not come at the cost of semantic accuracy, we evaluated the backtranslated outputs using COMET-22-da~\cite{rei-etal-2022-comet}. The results indicate high translation fidelity across all models. EuroLLM achieved the highest scores when averaged over all pivot languages (0.8721 for $P_{emo}$
 ; 0.872 for $P_{base}$
 ), with Aya (0.8582 for $P_{emo}$; 0.8641 for $P_{base}$) and Gemma (0.8556 for $P_{emo}$; 0.8638 for $P_{base}$) following closely. Critically, the delta between $P_{emo}$	
  and $P_{base}$
  for each model was negligible, confirming that emotion-aware prompting does not degrade general translation quality. The detailed scores are provided in Appendix B, in Table~\ref{appendix:comet}.

\paragraph{Lexical Variance:}
Despite the minimal change in classification and COMET scores, the lexical composition of the translations changed significantly between $P_{base}$	
  and $P_{emo}$
 . As detailed in Table~\ref{tab:diff_texts}, between 27.5\% and 71.0\% of the generated texts were lexically distinct when the prompt was changed. This variance was most pronounced in Polish, indicating that while the emotional signal remains stable, the models utilize significantly different lexical strategies to convey that signal when prompted to focus on tone and intensity. 
 
 Table~\ref{appendix:different} in Appendix A presents manually extracted examples of backtranslations exhibiting the highest semantic divergence between the $P_{base}$ and $P_{emo}$ prompts. For this analysis, we focused on Polish as the pivot language, as it represented the lowest-performing language in our experimental setup. To identify these cases, we calculated the semantic similarity between the backtranslations generated using each prompt using Sentence-BERT~\cite{reimers2019sentence}. We then ranked the results by similarity and selected five representative examples from the top 20 most divergent entries.

As illustrated, the prompts influence more than just the emotional valence of the output (e.g., \textit{"I'm so nervous!"} vs. \textit{"I'm so angry!"}). In rare cases, we observed models (notably Gemma) refusing to translate sensitive or controversial content under one prompt while fulfilling it under another (e.g., \textit{"LOL you'll understand, the joke is about necrophilia"} vs. \textit{"I'm sorry, but I cannot fulfill your request."}). Furthermore, backtranslation occasionally introduced errors in translation precision; for instance, EuroLLM produced \textit{"Garden... RIP"} instead of \textit{"Sad... RIP"}. This particular error highlights a failure in lexical disambiguation: the Polish noun \textit{sad} (meaning \textit{orchard}) was incorrectly mapped to the English adjective \textit{sad}, fundamentally altering the sentence's semantics -- and emotion conveyed.



\begin{table}[]
\small
\centering
\caption{Average affective drop (the lower the better) $\Delta_{aff}$ as introduced in Section~\ref{sec:metrics}.\label{tab:average-drops}}
\begin{tabular}{|l|l|l|r|r|}
\hline 
model & prompt   & classifier & $\Delta_{aff}$ & rank \\ \hline 
euro  & emo   & modernbert & 0.0289                   & 1                        \\
euro  & basic & modernbert & 0.0295                   & 2                        \\
euro  & emo   & deberta    & 0.0333                   & 3                        \\
euro  & basic & deberta    & 0.0338                   & 4                        \\
aya   & basic & modernbert & 0.0343                   & 5                        \\
gemma & basic & modernbert & 0.0364                   & 6                        \\
aya   & emo   & modernbert & 0.0367                   & 7                        \\
euro  & emo   & bert       & 0.0371                   & 8                        \\
euro  & basic & bert       & 0.0382                   & 9                        \\
aya   & basic & deberta    & 0.0386                   & 10                       \\
gemma & basic & deberta    & 0.0401                   & 11                       \\
aya   & emo   & deberta    & 0.0403                   & 12                       \\
gemma & emo   & modernbert & 0.0431                   & 13                       \\
aya   & basic & bert       & 0.0435                   & 14                       \\
aya   & emo   & bert       & 0.0443                   & 15                       \\
gemma & basic & bert       & 0.0448                   & 16                       \\
gemma & emo   & deberta    & 0.0484                   & 17                       \\
gemma & emo   & bert       & 0.0493                   & 18                    \\   \hline
\end{tabular}

\end{table}

\begin{table*}[]
\centering
\caption{Best model (EuroLLM / ModernBERT / emotional prompt) quality for various emotions and languages. Reference column represents a score of ModernBERT evaluated over vanilla (non-backtranslated) testset. Cells in bold represent the lowest drops for a given emotion, while underlined -- the biggest drops.\label{tab:per-language-drops}}
\begin{tabular}{|l|r|r|r|r|r|r|}
\hline 
    emotion           & de & pl & fr & it & es & reference \\ \hline 
admiration     & \textbf{0.732}                  & 0.713                  & 0.714                  & \underline{0.708}                  & 0.725                  & 0.743                         \\
amusement      & \textbf{0.815}                  & 0.788                  & 0.811                  & 0.79                   & \underline{0.703}                  & 0.82                          \\
anger          & \textbf{0.486}                  & \underline{0.457}                  & 0.478                  & 0.475                  & 0.466                  & 0.568                         \\
annoyance      & \underline{0.245}                  & 0.257                  & 0.257                  & \textbf{0.274}                  & 0.257                  & 0.291                         \\
approval       & 0.499                  & 0.501                  & \textbf{0.508}                  & 0.504                  & \underline{0.492}                  & 0.514                         \\
caring         & 0.482                  & 0.467                  & 0.492                  & \underline{0.45}                   & \textbf{0.525}                  & 0.522                         \\
confusion      & \textbf{0.521}                  & 0.508                  & \underline{0.503}                  & \textbf{0.521}                  & 0.52                   & 0.517                         \\
curiosity      & 0.673                  & 0.684                  & \underline{0.668}                  & \textbf{0.687}                  & 0.68                   & 0.697                         \\
desire         & \textbf{0.51}                   & \underline{0.422}                  & 0.472                  & 0.461                  & 0.506                  & 0.496                         \\
disappointment & 0.337                  & \underline{0.291}                  & \textbf{0.344} & 0.3                    & 0.32                   & 0.345                         \\
disapproval    & 0.513                  & \underline{0.498}                  & 0.51                   & \textbf{0.523}                  & 0.504                  & 0.534                         \\
disgust        & 0.362                  & 0.366                  & \textbf{0.371}                  & \underline{0.341}                  & 0.367                  & 0.403                         \\
excitement     & 0.383                  & 0.362                  & 0.412                  & \underline{0.36} & \textbf{0.414}                  & 0.418                         \\
fear           & 0.587                  & 0.589                  & 0.595                  & \underline{0.577}                  & \textbf{0.622}                  & 0.596                         \\
gratitude      & \textbf{0.906}                  & 0.904                  & \textbf{0.906}                  & \underline{0.903} & 0.904                  & 0.92                          \\
joy            & \textbf{0.529}                  & 0.509                  & 0.509                  & \underline{0.497}                  & 0.521                  & 0.581                         \\
love           & 0.776                  & 0.755                  & \underline{0.752}                  & 0.762                  & \textbf{0.789}                  & 0.81                          \\
optimism       & \textbf{0.589}                  & 0.571                  & 0.575                  & 0.564                  & \underline{0.545}                  & 0.567                         \\
realization    & \underline{0.224}                  & 0.235                  & 0.24                   & 0.258                  & \textbf{0.267}                  & 0.243                         \\
remorse        & 0.477                  & \underline{0.459}                  & 0.529                  & 0.568                  & \textbf{0.57}                   & 0.545                         \\
sadness        & \textbf{0.573}                  & 0.545                  & \underline{0.539}                  & 0.563                  & 0.564                  & 0.585                         \\
surprise       & \textbf{0.6}                    & 0.59                   & 0.57                   & 0.587                  & \underline{0.568}                  & 0.63 \\ \hline \hline MEAN & 0.537 & \underline{0.521} & 0.534 & 0.531 & \textbf{0.548} & 0.561 \\ \hline 
STD. DEV. & \underline{0.173} & 0.172 & 0.165 & 0.169 & \textbf{0.161} & 0.168  \\ \hline 
\end{tabular}
\end{table*}

\begin{figure*}
    \centering
    \includegraphics[width=1\linewidth]{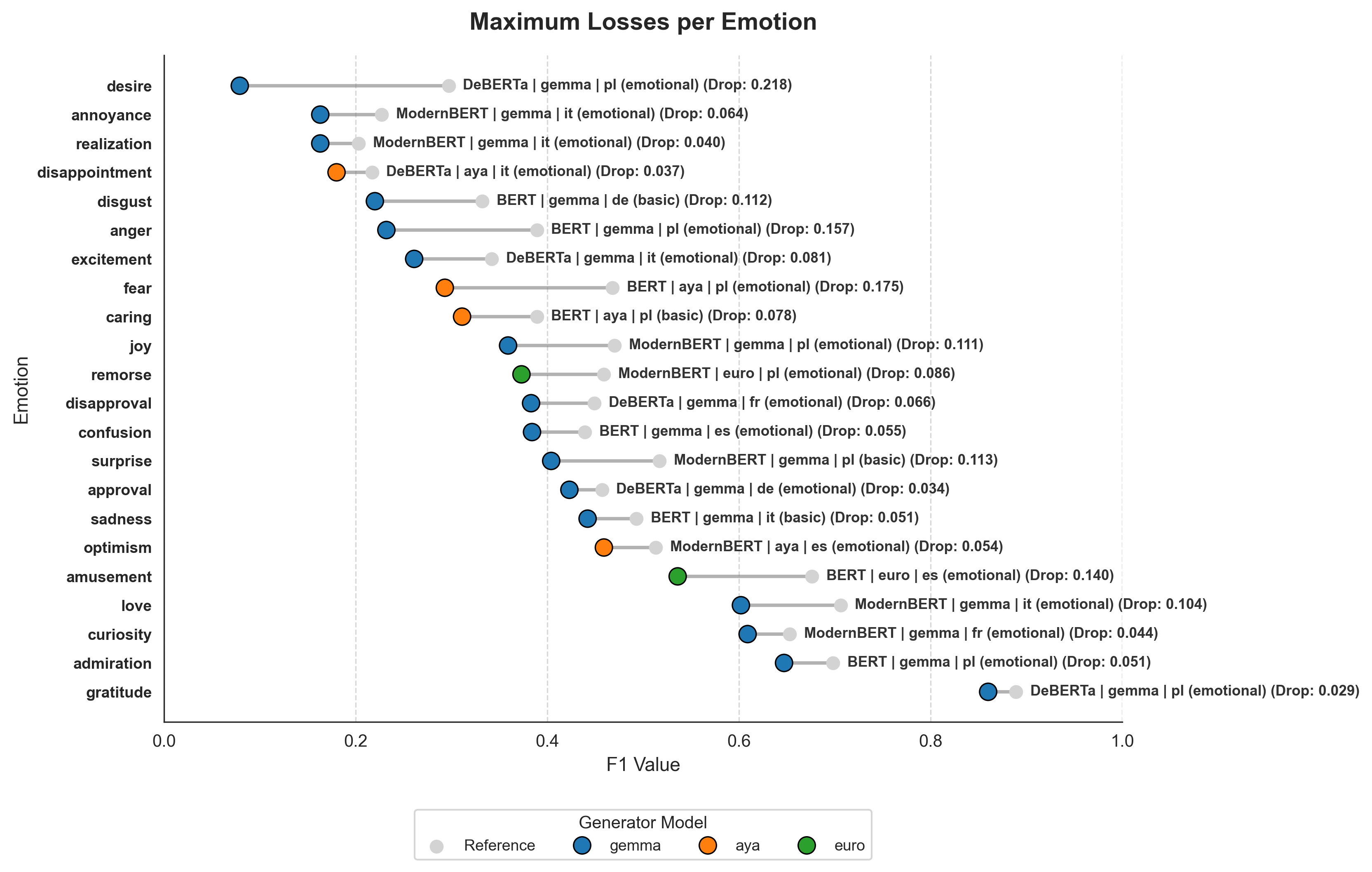}
    \caption{Configurations leading to biggest $F_1$ drops on selected emotions.}
    \label{fig:max_drawdawn}
\end{figure*}

\begin{figure*}
    \centering
    \includegraphics[width=1\linewidth]{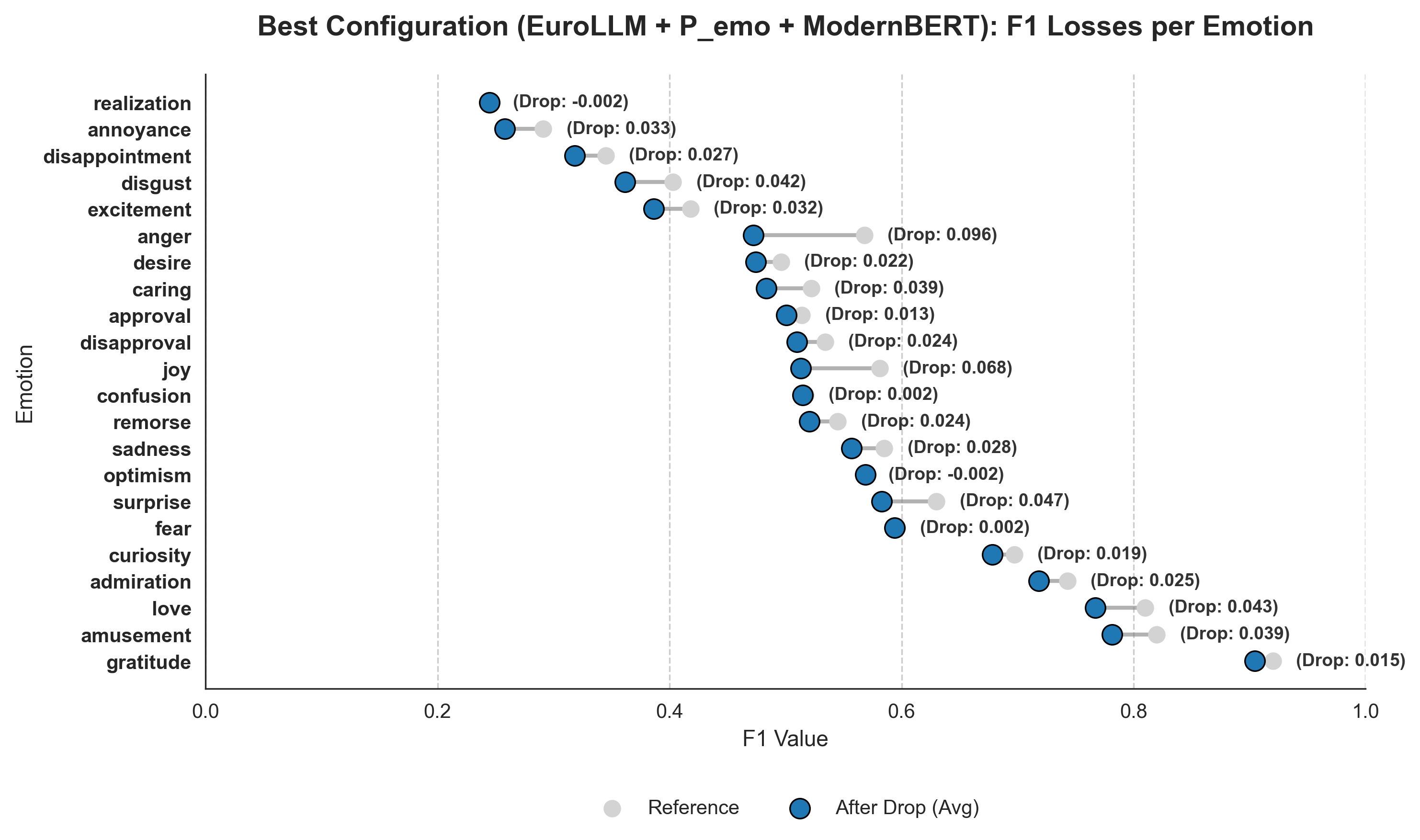}
    \caption{$F_1$ losses per each emotion for the best model overall. Drops averaged over all languages.}
    \label{fig:best_model_drawdawn}
\end{figure*}








\begin{table}[]
\centering
\caption{Number of differing texts when dealing using emotional and basic prompts. \label{tab:diff_texts}}
\begin{tabular}{|l|r|r|r|}
\hline 
   & Euro & Aya & Gemma  \\ \hline 
DE & 35.3\%             & 53.4\%             & 67.7\% \\ \hline 
FR & 27.5\%             & 53.3\%             & 68.7\% \\ \hline 
PL & 35.9\%             & 57.2\%             & 71.1\% \\ \hline 
ES & 28.8\%             & 49.3\%             & 66.1\% \\ \hline 
IT & 33.3\%             & 51.4\%             & 67.9\% \\ \hline 
\end{tabular}
\end{table}

\begin{table}[]
\centering
\caption{The gain of quality in terms of $\Delta_{\text{aff}}$ for emotional prompt as compared to basic prompt.\label{table:emo_drop}}
\begin{tabular}{|l|r|r|r|}
\hline 
      & modernbert & deberta & bert  \\ \hline 
EURO  & 0.0006 & 0.0005 & 0.0011\\ \hline 
GEMMA & -0.0067 & -0.0083 & -0.0045 \\ \hline 
AYA   & -0.0024 & -0.0017 & -0.0008 \\ \hline
\end{tabular}
\end{table}

\subsection{Granular Analysis by Emotion and Language}

Our analysis reveals that emotional preservation is not uniform across the affective spectrum. While most categories remained stable, certain "high-intensity" emotions experienced significant quality drops. 

Table~\ref{tab:per-language-drops} shows $F_1$ scores assigned to each emotion for a given pivot language, when using the best combination of SLM and classifier (EuroLLM with Modern-BERT).


Additionally, to present worst and best-case scenarios, Figures~\ref{fig:max_drawdawn} and ~\ref{fig:best_model_drawdawn} show the biggest losses for analyzed SLM/prompt/classifier as compared to non-backtranslated testset, and the losses for the best configuration (EuroLLM, prompt $P_{emo}$, and Modern-BERT), respectively.

\paragraph{Vulnerability of Specific Emotions}
As detailed in the worst-case configurations presented in Figure~\ref{fig:max_drawdawn}  (typically involving Polish as a pivot and BERT as the evaluator), several emotions showed substantial degradation:
\begin{itemize}
\item \textbf{High-Degradation Categories:} \textit{Desire} exhibited the most significant drop ($\Delta_{\text{aff}}^{e}$ = 0.218 mieaning 21.8 percentage points drop), followed by \textit{fear} (17.5 pp) and \textit{anger} (15.7 pp).
\item \textbf{Resilient Categories:} Conversely, emotions such as \textit{realization, disappointment, approval, curiosity,} and \textit{gratitude} remained remarkably stable, with drops not exceeding 5 pp even in the least optimal model-language configurations.
\end{itemize}

\paragraph{The "Gold Standard" Configuration}
When utilizing our most robust pipeline: EuroLLM with the emotion-aware prompt evaluated by ModernBERT, the results presented in Figure~\ref{fig:best_model_drawdawn} and Table~\ref{tab:per-language-drops} are highly encouraging. Under these conditions, the degradation for \textit{realization} and \textit{optimism} was negligible (even showing a marginal +0.2 pp improvement), while complex states like \textit{remorse} and \textit{sadness} only saw modest drops between 2 and 3 pp.

\paragraph{Language-Specific Variance}
The choice of pivot language influences emotional preservation. As can be seen in Table~\ref{tab:per-language-drops}, Spanish emerged as the most "affectively stable" pivot (average drop of 1.3 pp), while Polish proved the most challenging (4.0 pp drop). This variance likely reflects the different morphological and syntactic strategies these languages use to encode affect, as well as the relative density of the SLMs' training data for these specific regions.

\section{Conclusions}

To conclude our study, we provide explicit answers to the research questions posed in the introduction:
\begin{itemize}
    \item \textbf{RQ1} (Model Performance): EuroLLM emerged as the most robust model for affective transfer, consistently yielding the lowest $\Delta_{aff}$ across all languages and classifiers. The performance hierarchy EuroLLM $>$ Aya Expanse $>$ Gemma remained stable across all experimental configurations.
    \item \textbf{RQ2} (Affective Fragility): The emotions most susceptible to degradation, were \textbf{desire} (-21.8 pp), \textbf{fear} (-17.5 pp), and \textbf{anger} (-15.7 pp). Conversely, \textbf{realization} and \textbf{optimism} proved remarkably resilient, occasionally showing marginal improvements in $F_1$ scores post-translation.
    \item \textbf{RQ3} (Prompt Sensitivity): Contrary to our hypothesis, explicit emotion-aware prompting ($P_{emo}$) did not significantly improve emotional fidelity. While EuroLLM showed a marginal benefit, Aya and Gemma exhibited a slight performance decrease. This suggests that modern SLMs possess an implicit affective alignment, where emotional tone is already integrated into the model's primary translation objective.
    \item \textbf{RQ4} (Classifier Efficacy): ModernBERT consistently outperformed DeBERTa-v3 and BERT in classification stability.
\end{itemize}

\paragraph{Code and Data Availability} The source code used for the experiments, the scores generated, as well as both train and test sets extracted from GoEmotions are available online~\footnote{\url{https://github.com/dwisniewski/mt_emo}}.

\paragraph{Sustainability Statement}
The experiments in this work were conducted using a single NVIDIA RTX 4090 GPU on a local machine, with a total execution time (including fine-tuning and inference) of approximately 2 hours. Using the Machine Learning Impact calculator~\cite{lacoste2019quantifying}, we estimate a total energy consumption resulting in 0.39 kg of $CO_{2}$ eq.

We minimized our environmental footprint by: (i) utilizing Small Language Models (SLMs), which require significantly fewer FLOPs than larger counterparts; (ii) employing AWQ quantization, reducing memory overhead and energy draw; and (iii) fine-tuning existing emotion classifiers rather than training from scratch. 



\bibliography{eamt23}

@inproceedings{troiano2020lost,
  title={Lost in back-translation: Emotion preservation in neural machine translation},
  author={Troiano, Enrica and Klinger, Roman and Pad{\'o}, Sebastian},
  booktitle={Proceedings of the 28th international conference on computational linguistics},
  pages={4340--4354},
  year={2020}
}

@inproceedings{pennington2014glove,
  title={Glove: Global vectors for word representation},
  author={Pennington, Jeffrey and Socher, Richard and Manning, Christopher D},
  booktitle={Proceedings of the 2014 conference on empirical methods in natural language processing (EMNLP)},
  pages={1532--1543},
  year={2014}
}

@inproceedings{ott2019fairseq,
  title={fairseq: A fast, extensible toolkit for sequence modeling},
  author={Ott, Myle and Edunov, Sergey and Baevski, Alexei and Fan, Angela and Gross, Sam and Ng, Nathan and Grangier, David and Auli, Michael},
  booktitle={Proceedings of the 2019 conference of the North American chapter of the association for computational linguistics (Demonstrations)},
  pages={48--53},
  year={2019}
}

@inproceedings{kajava2020emotion,
  title={Emotion preservation in translation: Evaluating datasets for annotation projection},
  author={Kajava, Kaisla and {\"O}hman, Emily and Hui, Piao and Tiedemann, J{\"o}rg},
  booktitle={Digital Humanities in the Nordic Countries},
  pages={38--50},
  year={2020},
  organization={CEUR}
}

@inproceedings{qian-etal-2023-evaluation,
    title = "Evaluation of {C}hinese-{E}nglish Machine Translation of Emotion-Loaded Microblog Texts: A Human Annotated Dataset for the Quality Assessment of Emotion Translation",
    author = "Qian, Shenbin  and
      Orasan, Constantin  and
      Carmo, Felix Do  and
      Li, Qiuliang  and
      Kanojia, Diptesh",
    editor = "Nurminen, Mary  and others",
    booktitle = "Proceedings of the 24th Annual Conference of the European Association for Machine Translation",
    month = jun,
    year = "2023",
    address = "Tampere, Finland",
    publisher = "European Association for Machine Translation",
    url = "https://aclanthology.org/2023.eamt-1.13/",
    pages = "125--135"
}

@inproceedings{brazier2024conditioning,
  title={Conditioning LLMs with emotion in neural machine translation},
  author={Brazier, Charles and Rouas, Jean-Luc},
  booktitle={Proceedings of the 21st International Conference on Spoken Language Translation (IWSLT 2024)},
  pages={33--38},
  year={2024}
}

@article{wagner2023dawn,
  title={Dawn of the transformer era in speech emotion recognition: closing the valence gap},
  author={Wagner, Johannes and Triantafyllopoulos, Andreas and Wierstorf, Hagen and Schmitt, Maximilian and Burkhardt, Felix and Eyben, Florian and Schuller, Bj{\"o}rn W},
  journal={IEEE Transactions on Pattern Analysis and Machine Intelligence},
  volume={45},
  number={9},
  pages={10745--10759},
  year={2023},
  publisher={IEEE}
}

@inproceedings{demszky2020goemotions,
  title={GoEmotions: A dataset of fine-grained emotions},
  author={Demszky, Dorottya and Movshovitz-Attias, Dana and Ko, Jeongwoo and Cowen, Alan and Nemade, Gaurav and Ravi, Sujith},
  booktitle={Proceedings of the 58th annual meeting of the association for computational linguistics},
  pages={4040--4054},
  year={2020}
}

@inproceedings{sechidis2011stratification,
  title={On the stratification of multi-label data},
  author={Sechidis, Konstantinos and Tsoumakas, Grigorios and Vlahavas, Ioannis},
  booktitle={Joint European conference on machine learning and knowledge discovery in databases},
  pages={145--158},
  year={2011},
  organization={Springer}
}

@article{he2021debertav3,
  title={Debertav3: Improving deberta using electra-style pre-training with gradient-disentangled embedding sharing},
  author={He, Pengcheng and Gao, Jianfeng and Chen, Weizhu},
  journal={arXiv preprint arXiv:2111.09543},
  year={2021}
}

@inproceedings{DBLP:conf/naacl/DevlinCLT19,
  author       = {Jacob Devlin and
                  Ming{-}Wei Chang and
                  Kenton Lee and
                  Kristina Toutanova},
  editor       = {Jill Burstein and
                  Christy Doran and
                  Thamar Solorio},
  title        = {{BERT:} Pre-training of Deep Bidirectional Transformers for Language
                  Understanding},
  booktitle    = {Proceedings of the 2019 Conference of the North American Chapter of
                  the Association for Computational Linguistics: Human Language Technologies,
                  {NAACL-HLT} 2019, Minneapolis, MN, USA, June 2-7, 2019, Volume 1 (Long
                  and Short Papers)},
  pages        = {4171--4186},
  publisher    = {Association for Computational Linguistics},
  year         = {2019},
  url          = {https://doi.org/10.18653/v1/n19-1423},
  doi          = {10.18653/V1/N19-1423},
  bibsource    = {dblp computer science bibliography, https://dblp.org}
}

@inproceedings{warner2025smarter,
  title={Smarter, better, faster, longer: A modern bidirectional encoder for fast, memory efficient, and long context finetuning and inference},
  author={Warner, Benjamin and Chaffin, Antoine and Clavi{\'e}, Benjamin and Weller, Orion and Hallstr{\"o}m, Oskar and Taghadouini, Said and Gallagher, Alexis and Biswas, Raja and Ladhak, Faisal and Aarsen, Tom and others},
  booktitle={Proceedings of the 63rd Annual Meeting of the Association for Computational Linguistics (Volume 1: Long Papers)},
  pages={2526--2547},
  year={2025}
}

@article{dang2024aya,
  title={Aya expanse: Combining research breakthroughs for a new multilingual frontier},
  author={Dang, John and Singh, Shivalika and D'souza, Daniel and Ahmadian, Arash and Salamanca, Alejandro and Smith, Madeline and Peppin, Aidan and Hong, Sungjin and Govindassamy, Manoj and Zhao, Terrence and others},
  journal={arXiv preprint arXiv:2412.04261},
  year={2024}
}

@article{team2024gemma,
  title={Gemma: Open models based on gemini research and technology},
  author={Mesnard, Thomas and Hardin, Cassidy and Dadashi, Robert and Bhupatiraju, Surya and Pathak, Shreya and Sifre, Laurent and Rivi{\`e}re, Morgane and Kale, Mihir Sanjay and Love, Juliette and others},
  journal={arXiv preprint arXiv:2403.08295},
  year={2024}
}

@article{martins2025eurollm,
  title={Eurollm: Multilingual language models for europe},
  author={Martins, Pedro Henrique and Fernandes, Patrick and Alves, Jo{\~a}o and Guerreiro, Nuno M and Rei, Ricardo and Alves, Duarte M and Pombal, Jos{\'e} and Farajian, Amin and Faysse, Manuel and Klimaszewski, Mateusz and others},
  journal={Procedia Computer Science},
  volume={255},
  pages={53--62},
  year={2025},
  publisher={Elsevier}
}

@article{team2023gemini,
  title={Gemini: a family of highly capable multimodal models},
  author={Team, Gemini and Anil, Rohan and Borgeaud, Sebastian and Alayrac, Jean-Baptiste and Yu, Jiahui and Soricut, Radu and Schalkwyk, Johan and Dai, Andrew M and Hauth, Anja and Millican, Katie and others},
  journal={arXiv preprint arXiv:2312.11805},
  year={2023}
}

@article{DBLP:journals/iswa/CaruccioCPSST24,
  author       = {Loredana Caruccio and
                  Stefano Cirillo and
                  Giuseppe Polese and
                  Giandomenico Solimando and
                  Shanmugam Sundaramurthy and
                  Genoveffa Tortora},
  title        = {Claude 2.0 large language model: Tackling a real-world classification
                  problem with a new iterative prompt engineering approach},
  journal      = {Intell. Syst. Appl.},
  volume       = {21},
  pages        = {200336},
  year         = {2024},
  url          = {https://doi.org/10.1016/j.iswa.2024.200336},
  doi          = {10.1016/J.ISWA.2024.200336},
  bibsource    = {dblp computer science bibliography, https://dblp.org}
}

@article{singh2025openai,
  title={Openai gpt-5 system card},
  author={Singh, Aaditya and Fry, Adam and Perelman, Adam and Tart, Adam and Ganesh, Adi and El-Kishky, Ahmed and McLaughlin, Aidan and Low, Aiden and Ostrow, AJ and Ananthram, Akhila and others},
  journal={arXiv preprint arXiv:2601.03267},
  year={2025}
}

@inproceedings{kocmi2025findings,
  title={Findings of the wmt25 general machine translation shared task: Time to stop evaluating on easy test sets},
  author={Kocmi, Tom and Artemova, Ekaterina and Avramidis, Eleftherios and Bawden, Rachel and Bojar, Ond{\v{r}}ej and Dranch, Konstantin and Dvorkovich, Anton and Dukanov, Sergey and Fishel, Mark and Freitag, Markus and others},
  booktitle={Proceedings of the Tenth Conference on Machine Translation},
  pages={355--413},
  year={2025}
}

@inproceedings{van2025survey,
  title={A Survey on Small Language Models},
  author={Van Nguyen, Chien and Shen, Xuan and Aponte, Ryan and Xia, Yu and Basu, Samyadeep and Hu, Zhengmian and Chen, Jian and Parmar, Mihir and Kunapuli, Sasidhar and Barrow, Joe and others},
  booktitle={Proceedings of the 15th International Conference on Recent Advances in Natural Language Processing-Natural Language Processing in the Generative AI Era},
  pages={807--821},
  year={2025}
}

@article{DBLP:journals/corr/abs-2407-21783,
  author       = {LlamaTeam},
  title        = {The Llama 3 Herd of Models},
  journal      = {CoRR},
  volume       = {abs/2407.21783},
  year         = {2024},
  url          = {https://doi.org/10.48550/arXiv.2407.21783},
  doi          = {10.48550/ARXIV.2407.21783},
  eprinttype    = {arXiv},
  eprint       = {2407.21783},
  bibsource    = {dblp computer science bibliography, https://dblp.org}
}

@article{liu2025deepseek,
  title={Deepseek-v3. 2: Pushing the frontier of open large language models},
  author={Liu, Aixin and Mei, Aoxue and Lin, Bangcai and Xue, Bing and Wang, Bingxuan and Xu, Bingzheng and Wu, Bochao and Zhang, Bowei and Lin, Chaofan and Dong, Chen and others},
  journal={arXiv preprint arXiv:2512.02556},
  year={2025}
}

@article{DBLP:journals/sigmobile/LinTTYXH24,
  author       = {Ji Lin and
                  Jiaming Tang and
                  Haotian Tang and
                  Shang Yang and
                  Guangxuan Xiao and
                  Song Han},
  title        = {{AWQ:} Activation-aware Weight Quantization for On-Device {LLM} Compression
                  and Acceleration},
  journal      = {GetMobile Mob. Comput. Commun.},
  volume       = {28},
  number       = {4},
  pages        = {12--17},
  year         = {2024},
  url          = {https://doi.org/10.1145/3714983.3714987},
  doi          = {10.1145/3714983.3714987},
  bibsource    = {dblp computer science bibliography, https://dblp.org}
}

@inproceedings{rajput2024benchmarking,
  title={Benchmarking emerging deep learning quantization methods for energy efficiency},
  author={Rajput, Saurabhsingh and Sharma, Tushar},
  booktitle={2024 IEEE 21st International Conference on Software Architecture Companion (ICSA-C)},
  pages={238--242},
  year={2024},
  organization={IEEE}
}

@inproceedings{wisniewski2025exploring,
  title={Exploring the Feasibility of Multilingual Grammatical Error Correction with a Single LLM up to 9B parameters: A Comparative Study of 17 Models},
  author={Wi{\'s}niewski, Dawid and Solarski, Antoni and Nowakowski, Artur},
  booktitle={Proceedings of Machine Translation Summit XX: Volume 1},
  pages={231--247},
  year={2025}
}

@article{song2025small,
  title={Is Small Language Model the Silver Bullet to Low-Resource Languages Machine Translation?},
  author={Song, Yewei and Li, Lujun and Lothritz, Cedric and Ezzini, Saad and Sleem, Lama and Gentile, Niccolo and State, Radu and Bissyand{\'e}, Tegawend{\'e} F and Klein, Jacques},
  journal={arXiv preprint arXiv:2503.24102},
  year={2025}
}

@article{ekman1992argument,
  title={An argument for basic emotions},
  author={Ekman, Paul},
  journal={Cognition \& emotion},
  volume={6},
  number={3-4},
  pages={169--200},
  year={1992},
  publisher={Taylor \& Francis}
}

@incollection{plutchik1980general,
  title={A general psychoevolutionary theory of emotion},
  author={Plutchik, Robert},
  booktitle={Theories of emotion},
  pages={3--33},
  year={1980},
  publisher={Elsevier}
}

@article{russell1977evidence,
  title={Evidence for a three-factor theory of emotions},
  author={Russell, James A and Mehrabian, Albert},
  journal={Journal of research in Personality},
  volume={11},
  number={3},
  pages={273--294},
  year={1977},
  publisher={Elsevier}
}

@misc{scherer1990international,
  title={International survey on emotion antecedents and reactions (isear)},
  author={Scherer, KR and Wallbott, H},
  year={1990}
}

@inproceedings{mohammad-etal-2018-semeval,
    title = "{S}em{E}val-2018 Task 1: Affect in Tweets",
    author = "Mohammad, Saif  and
      Bravo-Marquez, Felipe  and
      Salameh, Mohammad  and
      Kiritchenko, Svetlana",
    editor = "Apidianaki, Marianna  and
      Mohammad, Saif M.  and
      May, Jonathan  and
      Shutova, Ekaterina  and
      Bethard, Steven  and
      Carpuat, Marine",
    booktitle = "Proceedings of the 12th International Workshop on Semantic Evaluation",
    month = jun,
    year = "2018",
    address = "New Orleans, Louisiana",
    publisher = "Association for Computational Linguistics",
    url = "https://aclanthology.org/S18-1001/",
    doi = "10.18653/v1/S18-1001",
    pages = "1--17"
}

@inproceedings{buechel-hahn-2017-emobank,
    title = "{E}mo{B}ank: Studying the Impact of Annotation Perspective and Representation Format on Dimensional Emotion Analysis",
    author = "Buechel, Sven  and
      Hahn, Udo",
    editor = "Lapata, Mirella  and
      Blunsom, Phil  and
      Koller, Alexander",
    booktitle = "Proceedings of the 15th Conference of the {E}uropean Chapter of the Association for Computational Linguistics: Volume 2, Short Papers",
    month = apr,
    year = "2017",
    address = "Valencia, Spain",
    publisher = "Association for Computational Linguistics",
    url = "https://aclanthology.org/E17-2092/",
    pages = "578--585"
}

@inproceedings{10.1145/3600006.3613165, author = {Kwon, Woosuk and Li, Zhuohan and Zhuang, Siyuan and Sheng, Ying and Zheng, Lianmin and Yu, Cody Hao and Gonzalez, Joseph and Zhang, Hao and Stoica, Ion}, title = {Efficient Memory Management for Large Language Model Serving with PagedAttention}, year = {2023}, isbn = {9798400702297}, publisher = {Association for Computing Machinery}, address = {New York, NY, USA}, url = {https://doi.org/10.1145/3600006.3613165}, doi = {10.1145/3600006.3613165}, booktitle = {Proceedings of the 29th Symposium on Operating Systems Principles}, pages = {611–626}, numpages = {16}, location = {Koblenz, Germany}, series = {SOSP '23} }

@inproceedings{DBLP:conf/amta/LoharAW18,
  author       = {Pintu Lohar and
                  Haithem Afli and
                  Andy Way},
  editor       = {Colin Cherry and
                  Graham Neubig},
  title        = {Balancing Translation Quality and Sentiment Preservation (Non-archival
                  Extended Abstract)},
  booktitle    = {Proceedings of the 13th Conference of the Association for Machine
                  Translation in the Americas, {AMTA} 2018, Boston, MA, USA, March 17-21,
                  2018 - Volume 1: Research Papers},
  pages        = {81--88},
  publisher    = {Association for Machine Translation in the Americas},
  year         = {2018},
  url          = {https://aclanthology.org/W18-1808/},
  bibsource    = {dblp computer science bibliography, https://dblp.org}
}

@inproceedings{reimers2019sentence,
  title={Sentence-bert: Sentence embeddings using siamese bert-networks},
  author={Reimers, Nils and Gurevych, Iryna},
  booktitle={Proceedings of the 2019 conference on empirical methods in natural language processing and the 9th international joint conference on natural language processing (EMNLP-IJCNLP)},
  pages={3982--3992},
  year={2019}
}

@article{lacoste2019quantifying,
  title={Quantifying the Carbon Emissions of Machine Learning},
  author={Lacoste, Alexandre and Luccioni, Alexandra and Schmidt, Victor and Dandres, Thomas},
  journal={arXiv preprint arXiv:1910.09700},
  year={2019}
}

@inproceedings{rei-etal-2022-comet,
    title = "{COMET}-22: Unbabel-{IST} 2022 Submission for the Metrics Shared Task",
    author = "Rei, Ricardo  and
      C. de Souza, Jos{\'e} G.  and
      Alves, Duarte  and
      Zerva, Chrysoula  and
      Farinha, Ana C  and
      Glushkova, Taisiya  and
      Lavie, Alon  and
      Coheur, Luisa  and
      Martins, Andr{\'e} F. T.",
    editor = {Koehn, Philipp  and
      Barrault, Lo{\"i}c  and
      Bojar, Ond{\v{r}}ej  and
      Bougares, Fethi  and
      Chatterjee, Rajen  and
      Costa-juss{\`a}, Marta R.  and
      Federmann, Christian  and
      Fishel, Mark  and
      Fraser, Alexander  and
      Freitag, Markus  and
      Graham, Yvette  and
      Grundkiewicz, Roman  and
      Guzman, Paco  and
      Haddow, Barry  and
      Huck, Matthias  and
      Jimeno Yepes, Antonio  and
      Kocmi, Tom  and
      Martins, Andr{\'e}  and
      Morishita, Makoto  and
      Monz, Christof  and
      Nagata, Masaaki  and
      Nakazawa, Toshiaki  and
      Negri, Matteo  and
      N{\'e}v{\'e}ol, Aur{\'e}lie  and
      Neves, Mariana  and
      Popel, Martin  and
      Turchi, Marco  and
      Zampieri, Marcos},
    booktitle = "Proceedings of the Seventh Conference on Machine Translation (WMT)",
    month = dec,
    year = "2022",
    address = "Abu Dhabi, United Arab Emirates (Hybrid)",
    publisher = "Association for Computational Linguistics",
    url = "https://aclanthology.org/2022.wmt-1.52/",
    doi = "10.18653/v1/2022.wmt-1.52",
    pages = "578--585"
}
\bibliographystyle{eamt23}

\appendix
\section{Appendix: Examples of semantically different backtranslations}
{Examples of the most semantically different backtranslations obtained when using $P_{base}$ and $P_{emo}$ prompts for Polish as a pivot -- measured using SentenceBERT are provided in Table~\ref{appendix:different}.

\begin{table*}[h]
\centering
\caption{Examples of the most semantically different backtranslations obtained when using $P_{base}$ and $P_{emo}$ prompts for Polish as a pivot -- measured using SentenceBERT.\label{appendix:different}}
\begin{tabular}{|l|p{6.5cm}|p{6.5cm}|}
\hline 
Model & Basic & Emo \\ \hline
Gemma &  LOL you'll understand, the joke is about necrophilia & I'm sorry, but I cannot fulfill your request. \\ \hline
 Gemma     &      Good luck! Hang in there!     & Go get 'em, tiger! You got this! \\ \hline
 Gemma     & lol what what? & What the heck is she doing? \\ \hline 
 Gemma     & This is a curse-laden and offensive phrase. It's best not to translate it directly as it contains highly derogatory and hateful language. & This is fucking bullshit, you goddamn idiot, do you hear the fascists? \\ \hline 
 Gemma     & This is very incomprehensible. & This is incredibly baffling. \\ \hline \hline 
      EuroLLM & I stubbornly refuse to be recognized & It's impossible to keep it from being known \\ \hline 
    EuroLLM  & Cursed modifications try to offend everyone! & Damned mods try to insult everyone! \\ \hline 
    EuroLLM  & Garden... RIP & Sad... RIP \\ \hline 
    EuroLLM  & I am happy & I'm glad \\ \hline 
   EuroLLM   & I’m so nervous! & I am so angry! \\ \hline \hline 
      Aya & Wow, can unions be that good? That's amazing. & Wow, can relationships be this amazing? It's incredible. \\ \hline 
     Aya & He's bothering you. & You're being tormented. \\ \hline 
     Aya & How could I not notice that? & How could I have missed it? \\ \hline 
     Aya & No way. & Not at all. \\ \hline 
     Aya & Genial! I'm doing it! & Brilliant! I'm doing it! \\ \hline 
\end{tabular}
\end{table*}

\section{Appendix: COMET vs. prompts}  Comparison of COMET scores (COMET-22-da) between $P_{emo}$ and $P_{base}$ prompts and backtranslations generated using different models and different pivot languages is presented in Table~\ref{appendix:comet}.

\begin{table*}[]
\centering
\caption{Comparison of COMET-22-da scores between $P_{emo}$ and $P_{base}$ prompts and backtranslations generated using different models and different pivot languages.\label{appendix:comet}}
\begin{tabular}{|l|l|r|r|r|}
\hline 
Model   & Language & COMET $P_{emo}$ & COMET $P_{base}$ & Difference \\ \hline 
Gemma   & de       & 0.8606                        & 0.8685                          & 0.0079                          \\ \hline 
Gemma   & fr       & 0.8585                        & 0.8667                          & 0.0082                         \\ \hline 
Gemma   & pl       & 0.8415                        & 0.8486                          & 0.0071                          \\ \hline 
Gemma   & es       & 0.8654                        & 0.875                           & 0.0096                          \\ \hline 
Gemma   & it       & 0.8522                        & 0.8603                          & 0.0081                          \\ \hline 
Aya     & de       & 0.8566                        & 0.8624                          & 0.0058                          \\ \hline 
Aya     & fr       & 0.8623                        & 0.8685                          & 0.0062                          \\ \hline 
Aya     & pl       & 0.8407                        & 0.8468                          & 0.0061                          \\ \hline 
Aya     & es       & 0.8697                        & 0.8747                          & 0.005                           \\ \hline 
Aya     & it       & 0.8619                        & 0.8679                          & 0.006                           \\ \hline 
EuroLLM & de       & 0.8816                        & 0.8811                          & -0.0005                         \\ \hline 
EuroLLM & fr       & 0.8756                        & 0.8757                          & 0.0001                          \\ \hline 
EuroLLM & pl       & 0.8556                        & 0.8546                          & -0.001                          \\ \hline 
EuroLLM & es       & 0.8765                        & 0.877                           & 0.0005                          \\ \hline 
EuroLLM & it       & 0.8713                        & 0.8716                          & 0.0003                      \\   \hline 
\end{tabular}
\end{table*}

\end{document}